\documentclass[11pt]{article}

\usepackage[preprint]{acl}

\usepackage{times}
\usepackage{latexsym}
\usepackage{microtype}
\usepackage{graphicx}
\usepackage{subcaption}
\usepackage{multirow}
\usepackage{booktabs} 
\usepackage{amsmath}
\usepackage{amssymb}
\usepackage{mathtools}
\usepackage{amsthm}
\usepackage[table]{xcolor}
\usepackage{makecell}
\usepackage{tcolorbox}
\usepackage[table]{xcolor}
\usepackage{enumitem}
\usepackage{algorithm}
\usepackage{tabularx}
\usepackage{algpseudocode} 
\usepackage{amsmath,amssymb}
\usepackage{courier}
\usepackage[table]{xcolor}
\usepackage{listings}
\usepackage{newfloat} 
\usepackage{listings}

\definecolor{codebg}{RGB}{248,248,248}
\definecolor{codeframe}{RGB}{210,210,210}
\definecolor{codetext}{RGB}{30,30,30}
\definecolor{codekw}{RGB}{0,92,160}
\definecolor{codecom}{RGB}{0,120,80}
\definecolor{codestr}{RGB}{140,60,0}

\lstdefinestyle{pycode}{
  language=Python,
  basicstyle=\ttfamily\footnotesize\color{codetext},
  keywordstyle=\bfseries\color{codekw},
  commentstyle=\itshape\color{codecom},
  stringstyle=\color{codestr},
  showstringspaces=false,
  breaklines=true,
  breakatwhitespace=true,
  tabsize=2,
  columns=fullflexible,
  keepspaces=true,
  frame=none,
  aboveskip=0pt,
  belowskip=0pt
}

\algrenewcommand\algorithmicrequire{\textbf{Inputs:}}
\algrenewcommand\algorithmicensure{\textbf{Outputs:}}

\newtcolorbox{fullwidthprompt}{
    colback=gray!5,        
    colframe=gray!60,      
    boxrule=0.5pt,        
    arc=2pt,               
    left=8pt, right=8pt, top=8pt, bottom=8pt, 
    fontupper=\small, 
    boxsep=0pt
}

\usepackage[T1]{fontenc}

\usepackage[utf8]{inputenc}

\usepackage{microtype}

\usepackage{inconsolata}

\usepackage{graphicx}

\title{Rethinking LLM-Driven Heuristic Design: Generating Efficient and Specialized Solvers via Dynamics-Aware Optimization}

\author{%
\normalfont\mdseries
Rongzheng Wang\textsuperscript{1},
Yihong Huang\textsuperscript{1},
Muquan Li\textsuperscript{1},
Jiakai Li\textsuperscript{1},\\
Di Liang\textsuperscript{2},
Bob Simons\textsuperscript{2},
Pei Ke\textsuperscript{1},
Shuang Liang\textsuperscript{1*},
Ke Qin\textsuperscript{1}
\\[0.6ex]
\textsuperscript{1} University of Electronic Science and Technology of China\\
\textsuperscript{2} Tencent Hunyuan\\[0.6ex]
\texttt{wangrongzheng@std.uestc.edu.cn}\quad \texttt{shuangliang@uestc.edu.cn}
}

\begin{document}

\maketitle
\begin{abstract}

Large Language Models (LLMs) have advanced the field of Combinatorial Optimization through automated heuristic generation. Instead of relying on manual design, this LLM-Driven Heuristic Design (LHD) process leverages LLMs to iteratively generate and refine solvers to achieve high performance. However, existing LHD frameworks face two critical limitations: (1) Endpoint-only evaluation, which ranks solvers solely by final gap to a reference solution, ignoring the convergence process and runtime efficiency; (2) High adaptation costs, where distribution shifts necessitate re-adaptation to generate specialized solvers for heterogeneous instance groups. To address these issues, we propose Dynamics-Aware Solver Heuristics (DASH), a framework that co-optimizes solver search mechanisms and runtime schedules guided by a convergence-aware metric, thereby identifying efficient and high-performance solvers. Furthermore, to mitigate expensive re-adaptation, DASH incorporates Profiled Library Retrieval (PLR), which maintains group-specialized solvers for profile-aware warm starts. These solvers are archived concurrently during evolution, allowing DASH to reuse matched specialists across heterogeneous distributions without restarting adaptation.
Experiments on four combinatorial optimization problems demonstrate that DASH improves runtime efficiency by over 4$\times$ while outperforming prior LHD baselines in the overall balance between gap and runtime across diverse problem scales.
Furthermore, by enabling profile-aware warm starts, DASH maintains lower gap under distribution shift while reducing LLM adaptation costs by about 90\%.
\end{abstract}

\section{Introduction}

Many fundamental problems in computing and engineering, including routing, scheduling, and chip placement, are combinatorial optimization problems~\cite{korte2008combinatorial}. In practice, high-performance solvers for these problems rely on carefully hand-crafted heuristics~\cite{burke2013hyper}. Due to the extensive search space, executing such heuristic solvers is computationally expensive. Beyond the high design cost, these heuristics often struggle with generalization: when the instance distribution changes (e.g., in size or density), sustaining performance typically requires re-design for re-adaptation~\cite{wolpert1997nofreelunch}.

LLM-Driven Heuristic Design (LHD)~\cite{yao2025multi, wu2025efficient} alleviates this burden by automatically generating and improving solvers. Recent works have advanced this domain by establishing a fundamental iterative workflow in which LLMs generate candidate solvers, evaluate them through execution feedback, and select promising ones for further refinement (e.g., FunSearch~\cite{romera2024funsearch}, EoH~\cite{DBLP:conf/icml/0044TY0LWL024}, and ReEvo~\cite{DBLP:conf/nips/Ye0CBHKPS24}). However, the evaluation process requires repeatedly generating and executing time-consuming solvers, meaning that practical LHD still faces two key challenges.


\begin{figure*}[t]
    \centering
    \begin{subfigure}[c]{0.44\linewidth} 
        \centering
        \includegraphics[width=\linewidth]{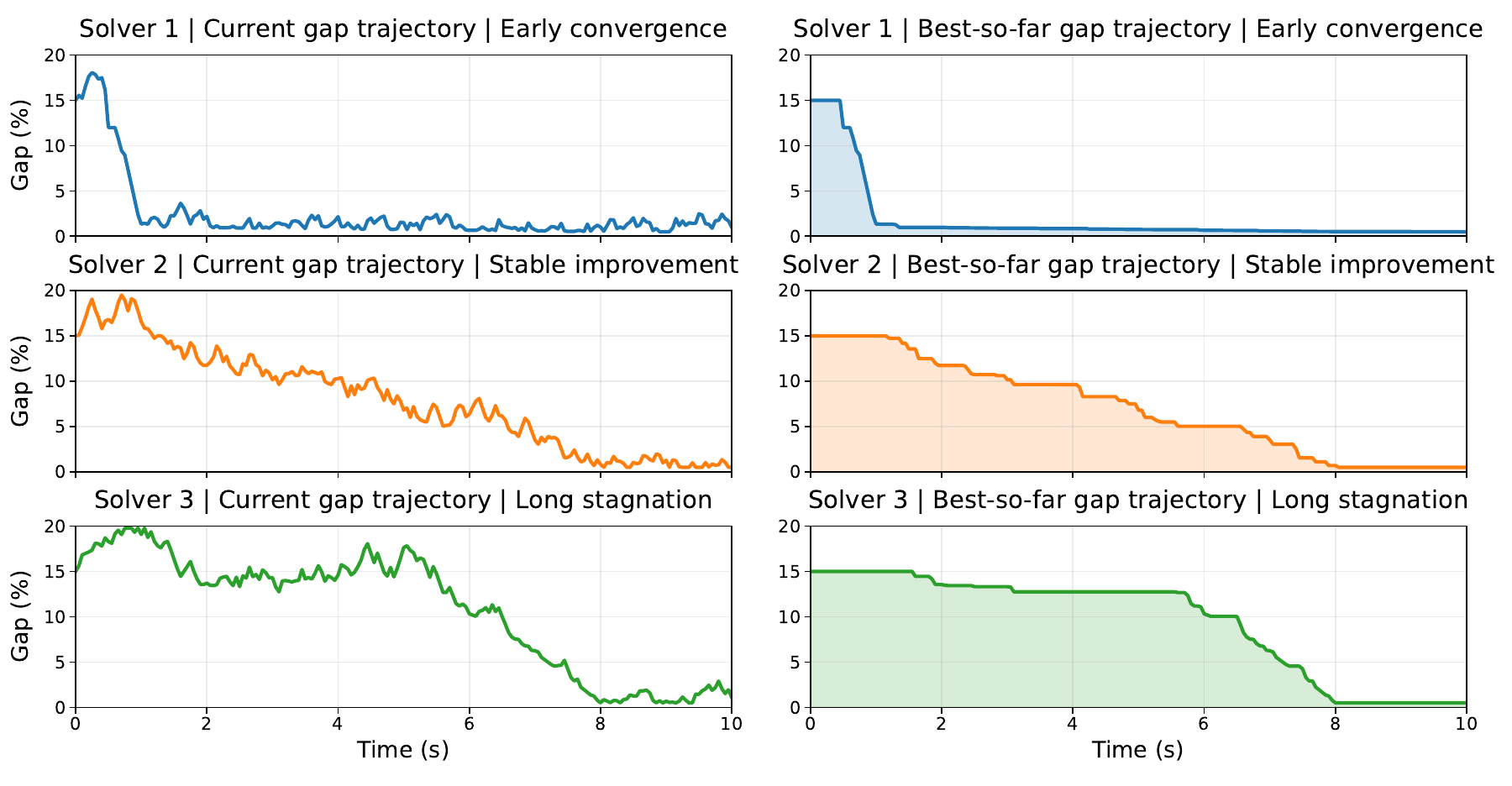}
    \end{subfigure}
    \hfill
    \begin{subfigure}[c]{0.54\linewidth}
        \centering
        \includegraphics[width=\linewidth]{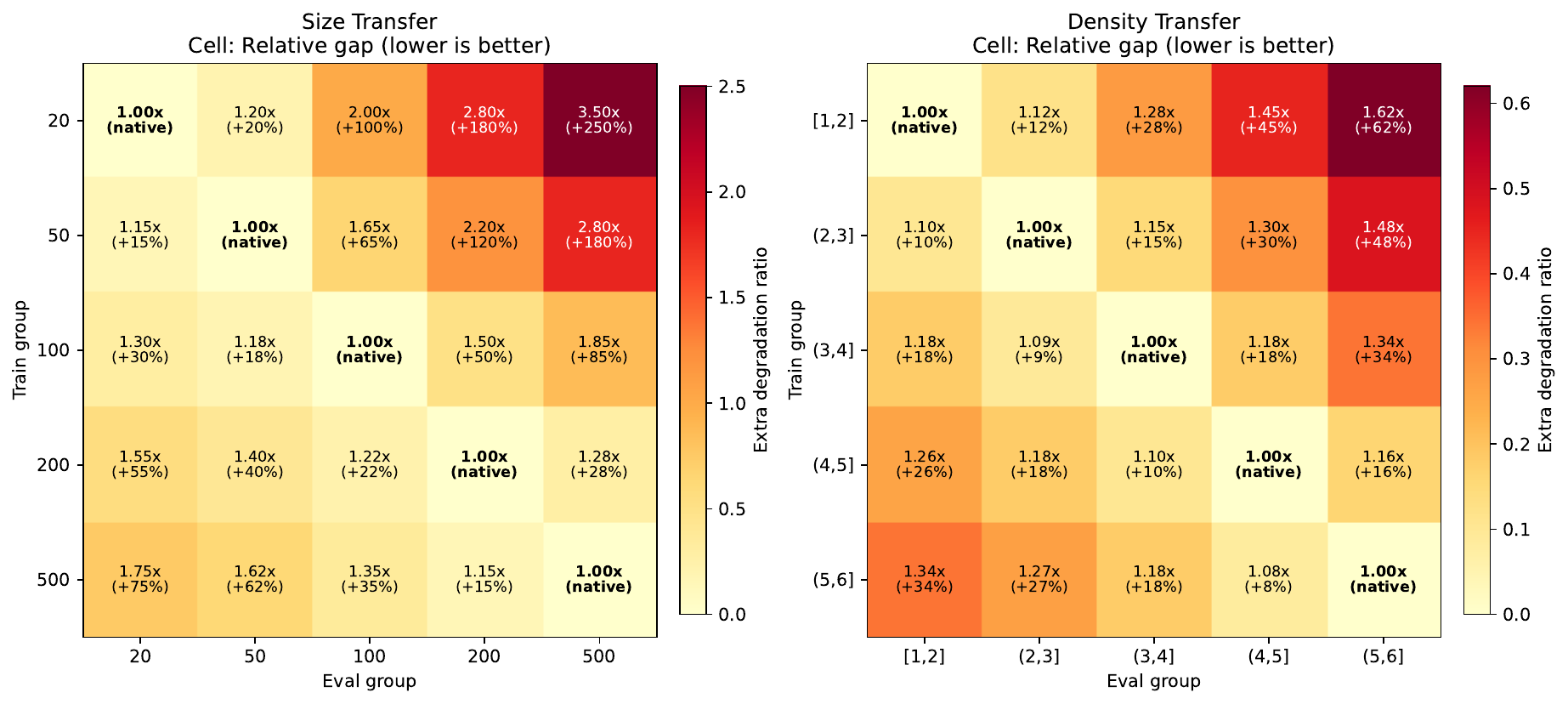}
    \end{subfigure}

    \caption{\textbf{Motivating Experiments on TSP tasks using Guided Local Search~\cite{voudouris1999gls} as solver backbone and under a 10s time limit.}
    (\emph{Left}) Three solvers reach similar final gaps despite different trajectories, rendering them nearly indistinguishable under endpoint-only evaluation. Solver 1 converges earlier, achieving lower gaps at earlier times and thereby leaving more of the time budget for subsequent improvements.
    (\emph{Right}) Performance significantly degrades when solvers are transferred across shifts in problem size (node count) or density (node clustering), further underscoring the necessity of generating specialized solvers for distinct instance distributions.}
    \label{icml-historical}
    
\end{figure*}

\paragraph{Challenge 1: Generating efficient solvers}
Most LHD frameworks~\cite{zheng2025mctsahd, dat2025hsevo} select candidate solvers only by the final score (e.g., gap to an optimal or best-known value), thereby ignoring a key dimension: the convergence trajectory within the time. Crucially, early convergence often indicates the potential to further reduce the gap to the optimal~\cite{hansen1996monitoring}, yet comparable final score makes it difficult to distinguish such efficient solvers. Some recent efficiency-aware LHD works have started to incorporate time objectives in solver selection (e.g., MEoH~\cite{yao2025multi}). Nevertheless, relying solely on such efficiency signal fails to capture the convergence dynamics (e.g., early convergence, stable improvement, and long stagnation). As Figure~\ref{icml-historical} (left) shows, three solvers reach comparable final gaps with different trajectories, rendering them indistinguishable under endpoint-only evaluation.

\paragraph{Challenge 2: Generating specialized solvers}
LHD is costly as it requires repeatedly refining solvers on training instances~\cite{guo-etal-2025-nested}. Some methods reduce this cost by reusing information from prior evaluations to filter redundant candidates, thereby cutting down solver runs (e.g., Hercules~\cite{wu2025efficient}). However, real-world instances are heterogeneous: a solver that works well on one instance group may degrade under distribution shifts in scale or other instance characteristics~\cite{rice1976algorithm}. 
Consequently, maintaining performance across diverse groups requires group-wise adaptation, leading to repeated LLM-driven iteration. 
Although prior work lowers cost per iteration, it remains inefficient under distribution shifts. As Figure~\ref{icml-historical} (right) illustrates, solvers optimized for one instance group can degrade significantly on another, resulting in costly re-adaptation.

These observations underscore the necessity of assessing a solver by the temporal evolution of its solution quality, framing execution as a dynamical process. Accordingly, we propose Dynamics-Aware Solver Heuristics (\textbf{DASH}), a framework that improves solver performance and efficiency by optimizing how the solver searches and how it spends runtime. Under this view, we decompose solver design into the search mechanism (e.g., update rules and guidance design) and the runtime schedule (e.g., time allocation across phases). To quantify these dynamics, we introduce the Trajectory-aware Lyapunov Decay Rate (\textbf{tLDR})~\cite{khalil2002nonlinear}, which measures the rate and consistency of convergence and guides the co-evolution of search mechanism and runtime schedule in DASH.

To mitigate re-adaptation costs, DASH further incorporates Profiled Library Retrieval (\textbf{PLR}). PLR decouples archiving from evolution: within a single evolutionary process, it archives group-specific solvers, while continuing to evolve the global solver based on average performance. 
This strategy builds a diverse library within a single search and enables cost-effective profile-aware warm starts.
We validate DASH on four combinatorial optimization problems and demonstrate its generalizability across solver backbones. Results show that DASH improves runtime efficiency by over 4$\times$ while outperforming prior LHD baselines in the overall balance between gap and runtime. By enabling profile-aware warm starts, DASH maintains lower gap under distribution shift while reducing LLM adaptation costs by about 90\%.

Our contributions can be summarized as follows: 
\begin{itemize}[noitemsep,topsep=0pt]
\item We introduce the tLDR, a trajectory-aware metric that shifts evaluation from static endpoints to dynamic convergence efficiency, prioritizing fast and stable solvers. 
\item We propose DASH, a framework that co-evolves search mechanisms and runtime schedules. It incorporates Profiled Library Retrieval (PLR) to maintain specialized solvers for cost-effective profile-aware warm starts.
\item We validate DASH on four combinatorial optimization problems. Results show that it improves runtime efficiency by over 4$\times$ while outperforming LHD baselines in the overall balance between gap and runtime, and substantially reduces adaptation cost under distribution shift.
\end{itemize}

\section{Related Work}
\textbf{Real-World Combinatorial Optimization.} Combinatorial optimization plays a central role in real-world decision systems~\cite{korte2008combinatorial}, where routing, dispatching, scheduling, and resource allocation must be performed under strict operational constraints. In such settings, optimization is valuable not only because it improves solution quality, but also because it directly affects service efficiency, resource utilization, and system responsiveness. At the same time, practical deployments are complicated by dynamic arrivals, limited runtime budgets, and heterogeneous instance distributions. These challenges have been widely studied across a range of real-world applications, including route guidance and taxi services~\cite{YuanZXS13,YuanZZX13}, constrained route queries~\cite{LiYM13}, large-scale dispatching~\cite{TongSXLQT23}, airport ground handling~\cite{ZhouWCSZC23}, ride-hailing control~\cite{ZhangYYMW24}, and dynamic routing~\cite{ZhangYCL25}.
Real-world optimization is inherently context-dependent, and therefore calls for adaptive methods that can respond to varying instances and runtime constraints rather than relying on a single fixed solver.

\textbf{Heuristics and Solver Adaptation.} Classical combinatorial optimization has long relied on problem-specific heuristics and metaheuristics~\cite{korte2008combinatorial}, whose effectiveness often depends on carefully designed rules tailored to a solver and a problem family. Hyper-heuristics raise this level of generality by seeking to automate the selection or generation of heuristics across problem classes~\cite{burke2013hyper}. However, they still face persistent challenges: heuristic quality can vary substantially across instances, and good decisions often depend on available runtime as well as final solution quality. This has motivated a related research direction on instance-aware solver adaptation, including algorithm selection~\cite{rice1976algorithm,LindauerHHS15} and runtime-aware control inspired by anytime optimization~\cite{hansen1996monitoring,zilberstein1996anytime}, where the preferred solver or configuration changes with instance characteristics and budget regimes. These limitations have motivated recent LLM-driven approaches that attempt to automate heuristic discovery while retaining the flexibility needed for heterogeneous optimization scenarios.

\textbf{LLM-Driven Heuristic Design.}
Combinatorial optimization problems typically rely on heuristic solvers~\cite{garey1983computers}.
Recently, LLMs have advanced LLM-Driven Heuristic Design (LHD) by establishing an iterative generate, evaluate, and select workflow to synthesize and refine solvers~\cite{guo-etal-2025-nested, zheng2025mctsahd, dat2025hsevo}.
Notable frameworks include FunSearch~\cite{romera2024funsearch}, which couples program mutation with evolutionary search based on execution feedback; EoH~\cite{DBLP:conf/icml/0044TY0LWL024}, which co-evolves natural language ideas and code implementations; and ReEvo~\cite{DBLP:conf/nips/Ye0CBHKPS24}, which employs reflective prompts to guide the search.
However, most LHD methods select solvers solely based on endpoint metrics, ignoring the convergence process.
Although some efficiency-aware variants incorporate runtime objectives, they typically reduce the dynamic trajectory to a few aggregated statistics~\cite{yao2025multi}.
A complementary direction addresses evaluation costs by reusing mechanisms or predictor-based pruning to filter candidates~\cite{wu2025efficient,guo-etal-2025-nested}.
Although effective for reducing overhead, such pruning may limit search diversity and overlook promising solvers.
DASH addresses these limitations by optimizing the full convergence trajectory for efficiency and employing PLR for cost-effective adaptation. This allows archived specialists to be reused across heterogeneous instance profiles without re-evolving a separate solver for each profile group.

\section{Method}
\subsection{Solver Runs as Time-Evolving Trajectory}\label{sec:dynamics_view}


We adopt a dynamical systems view of solver execution under a time limit $T$.
For an instance $x$, a solver $\pi=(\theta,\sigma)$ induces a time-evolving solution trajectory $z(\tau)\in\mathcal{Z}(x)$,
where $\theta$ encodes the search mechanism (e.g., update rules and guidance design) and $\sigma$ encodes the runtime schedule
(e.g., time allocation across phases).
In practice, solvers update their solutions in discrete steps, so we denote by $z_k := z(\tau_k)$ the solver state at cumulative runtime $\tau_k$, and consider:
\begin{equation}
z_{k+1} = F(z_k; x,\theta,\sigma), \quad \tau_{k+1} = \tau_k + \Delta \tau_k,
\end{equation}
where $F(\cdot; x,\theta,\sigma)$ denotes the state transition induced by solver $\pi=(\theta,\sigma)$ on instance $x$, and $\Delta \tau_k$ is the measured time cost of step $k$.

\subsection{Lyapunov Potential and Incumbent Trajectory}\label{sec:lyapunov_potential}

\paragraph{Lyapunov potential definition.} To compare such trajectories across runs, we require a progress signal that measures the distance to a target solution $z^\star$ (e.g., an optimal or best-known solution) at each time.
In dynamical systems, Lyapunov functions formalize this idea as a generalized energy that is minimized at the target state~\cite{khalil2002nonlinear}. Adopting this perspective, we define a proxy potential $V:\mathcal{Z}(x)\rightarrow\mathbb{R}_{\ge 0}$ satisfying:
\begin{equation}
V(z^\star)=0,\qquad V(z)>0 \ \text{for}\ z\neq z^\star.
\label{eq:lyapunov_def}
\end{equation}

\paragraph{Incumbent trajectory.}
For a minimization objective $f_x(\cdot)$ with reference optimum $f^\star$ (exact or best-known),
we measure the distance to optimality via the relative gap:
$\mathrm{gap}_x(\tau)=\big(f_x(z(\tau)) - f_x^\star\big)/|f_x^\star|$.
Consistent with Eq.~\eqref{eq:lyapunov_def}, we set the time-dependent residual as the gap itself:
\begin{equation}
V(\tau) = \mathrm{gap}_x(\tau),
\label{eq:V_def}
\end{equation}
where $V(\tau)$ is short for $V(z(\tau))$.
Since heuristic search trajectories are typically stochastic and non-monotone, we extract persistent progress via the incumbent (best-so-far) trajectory:
\begin{equation}
V_{\mathrm{best}}(\tau)=\min_{0\le u\le \tau} V(u).
\label{eq:Vbest_def}
\end{equation}
By construction, $V_{\mathrm{best}}(\tau)$ is non-increasing, providing a monotone progress signal that aligns with an energy-descent view of optimization dynamics.

We then project the incumbent trajectory into logarithmic space to obtain a scale-consistent notion of progress.
In particular, many improvements in combinatorial optimization are naturally compared in relative terms (i.e., multiplicative reductions of the gap). Working in log space turns such multiplicative changes into additive decreases, making progress comparable across different residual magnitudes.
To ensure well-defined values even when the optimum is reached (i.e., $V_{\mathrm{best}}=0$), we employ a numerical lower bound $\delta > 0$:
\begin{equation}
\ell(\tau)=\ln \max\big(V_{\mathrm{best}}(\tau),\, \delta\big).
\label{eq:logVbest_def}
\end{equation}
We use $\ell(T)$ as the corresponding terminal log-residual under the time limit $T$.

\begin{table*}[t]
\centering
\small
\setlength{\tabcolsep}{6pt}
\renewcommand{\arraystretch}{1.15}
\caption{Search Mechanism ($\theta$) and Runtime Schedule ($\sigma$) in GLS for TSP.}
\label{tab:gls_variants_corrected_tkde}
\begin{tabularx}{\textwidth}{l|X|X}
\toprule
\textbf{Component}
& \textbf{Search Mechanism $\theta$}
& \textbf{Runtime Schedule $\sigma$} \\
\midrule
Initialization
& tour construction (fixed for fairness)
& (none; fixed single start) \\

Neighborhood
& candidate restriction and move neighborhood definition
& when to update / switch \\

Local improvement
& move operators and acceptance logic
& phase budgets and iteration/stop caps \\

Guidance
& guidance/penalty rule and how it interacts with search
& when to activate and how often to update \\

Perturbation
& perturbation operator
& triggering/frequency and (optional) intensity \\

Stopping
& (none)
& time cap and stagnation-based stopping \\
\bottomrule
\end{tabularx}
\end{table*}

\subsection{Trajectory-aware Lyapunov Decay Rate}\label{sec:tldr}

Previous LHD methods based solely on the terminal gap at time $T$ fail to capture optimization dynamics:
two solvers may achieve similar $\ell(T)$ while exhibiting different early progress and different persistence in low-residual regimes.
To summarize the incumbent evolution over the entire process, we compute the time-averaged log-residual:
\begin{equation}
J(T)=\frac{1}{T}\int_{0}^{T}\ell(\tau)\,d\tau.
\label{eq:JT_def}
\end{equation}
A smaller $J(T)$ implies that the solver stays in low-residual states for a larger fraction of time, reflecting both early improvement and sustained convergence.

To convert this trajectory-aggregated quantity into an interpretable decay rate, we derive an effective slope.
Specifically, we define an equivalent linear trajectory $\tilde{\ell}(\tau) = \ell(0) - k\tau$, anchored at the initial residual $\ell(0)$. 
We determine the decay slope $k$ by requiring that this hypothetical linear path yields the same time-averaged value $J(T)$ as the actual observed run over $[0,T]$.

The integral of this linear path is $T\ell(0) - \frac{1}{2}kT^2$. By equating this to observed integral $T J(T)$ and solving for $k$, we define the \textbf{Trajectory-aware Lyapunov Decay Rate (tLDR)}:
\begin{equation}
\mathrm{tLDR}(T) = \frac{2}{T}\big(\ell(0)-J(T)\big).
\label{eq:tldr_def_main}
\end{equation}
$\mathrm{tLDR}(T)$ is the effective decay slope of the incumbent log-residual trajectory over the interval $[0,T]$ induced by the time-averaged log-residual $J(T)$.
Equivalently, it is the constant slope of a linear surrogate $\tilde{\ell}(\tau)=\ell(0)-k\tau$ whose time-average matches the overall run.
A larger $\mathrm{tLDR}$ indicates faster and more sustained reduction in log-residual throughout the entire interval $[0,T]$, rather than progress concentrated only near the end.
For fair comparison, $\ell(0)$ is always measured from the same initial state for each instance, reused across solver evaluations.


In practice, tLDR is computed from logged incumbent traces by averaging the incumbent log-residual over wall-clock time and converting that average into an equivalent constant decay slope.
Concretely, given a logged incumbent trace $\{(\tau_j,z_j)\}_{j=0}^{m}$ with $0=\tau_0<\cdots<\tau_m=T$ and $\ell_j=\ell(\tau_j)=\ln\max\big(V(z_j),\delta\big)$, we use the following piecewise-constant form:

\begin{equation}
J(T)=\frac{1}{T}\sum_{j=0}^{m-1}(\tau_{j+1}-\tau_j)\ell_j,
\label{eq:tldr_discrete_avg}
\end{equation}
which gives the corresponding discrete form of tLDR:
\begin{equation}
\mathrm{tLDR}(T)=\frac{2}{T}\big(\ell_0-J(T)\big).
\label{eq:tldr_discrete_est}
\end{equation}

\subsection{Three Iteration Layers}\label{sec:three_layers}

The performance of a solver is governed by two interacting components: the mechanism $\theta$ and the schedule $\sigma$.
However, directly optimizing $\pi=(\theta,\sigma)$ is challenging due to their complex coupling.
To address this, we decompose optimization into three sequential layers. Across all layers, we employ a unified protocol based on the terminal log-residual $\ell(T)$, trajectory efficiency $\mathrm{tLDR}(T)$, and (for runtime schedule optimization) solver runtime $t_{\mathrm{run}}$.

However, heuristic search trajectories can be stochastic and highly dependent on specific instance characteristics. To obtain a stable evaluation signal, candidates are evaluated on a sampled batch $\mathcal{B}$. 
Acceptance decisions are based on the batch-averaged metrics: terminal log-residual $\bar{\ell}$, trajectory efficiency $\bar{k}$ (mean $\mathrm{tLDR}(T)$), and runtime $\bar{t}$. Lower is better for $\bar{\ell}$ and $\bar{t}$, whereas higher is better for $\bar{k}$. Each candidate is compared only with its direct parent on the same evaluation batch. We use a shared comparison margin $\epsilon$ across all three layers. A metric $x'$ is treated as being within the comparison margin of its parent value $x$ if:

\begin{equation}
\begin{aligned}
&|x' - x| \le \epsilon \quad \text{for } x=\bar{\ell}, \\
&|x' - x| \le \epsilon \cdot |x| \quad \text{for } x\in\{\bar{k},\bar{t}\}.
\label{eq:tol_def}
\end{aligned}
\end{equation}
Since $\bar{\ell}$ is defined in log space, absolute differences already correspond to relative changes in the underlying residual, whereas $\bar{k}$ and $\bar{t}$ are compared on their original scales using parent-relative margins.

\subsubsection{Mechanism Discovery Layer (MDL)}\label{sec:mdl}

MDL updates the mechanism $\theta$ while keeping the schedule $\sigma$ fixed.
In each iteration, the LLM produces a candidate mechanism $\theta'$ by editing the code of $\theta$ based on the parent and its evaluation feedback.
We then evaluate $\theta'$ under a hierarchical selection criterion relative to its parent. A candidate is accepted if it achieves a lower batch-averaged terminal log-residual $\bar{\ell}$; when $\bar{\ell}$ remains within the comparison margin of the parent, we accept the candidate only if it also achieves a sufficiently larger batch-averaged trajectory efficiency $\bar{k}$ according to the parent-relative criterion.

\begin{equation}
\begin{aligned}
&\mathrm{Accept}(\theta') \iff\;
\big(\bar{\ell}' \le \bar{\ell} - \epsilon\big) \\
&\lor\ \big(|\bar{\ell}'-\bar{\ell}|\le \epsilon \ \land\  \bar{k}' \ge (1+\epsilon)\bar{k}\big).
\end{aligned}
\label{eq:mdl_rule}
\end{equation}

\subsubsection{Mechanism Consolidation Layer (MCL)}\label{sec:mcl}

MCL controls the structural complexity of the evolving mechanism $\theta$.
Iterative evolution can lead to code bloat, where the mechanism accumulates brittle logic that overfits to specific instances.
In this layer, the LLM refactors the mechanism by rewriting the code structure (e.g., merging duplicated branches and removing dead or redundant logic) while keeping its intended behavior. We employ a preservation criterion: a consolidated candidate $\theta'$ is accepted only if both the batch-averaged terminal log-residual $\bar{\ell}$ and the batch-averaged trajectory efficiency $\bar{k}$ remain within the comparison margin of the parent.

\begin{equation}
\mathrm{Accept}(\theta') \iff
\big(|\bar{\ell}'-\bar{\ell}|\le \epsilon\big)
\ \land\
\big(|\bar{k}'-\bar{k}|\le \epsilon\cdot|\bar{k}|\big).
\label{eq:mcl_rule}
\end{equation}

\begin{figure*}[t]
    \centering \centerline{\includegraphics[width=\linewidth]{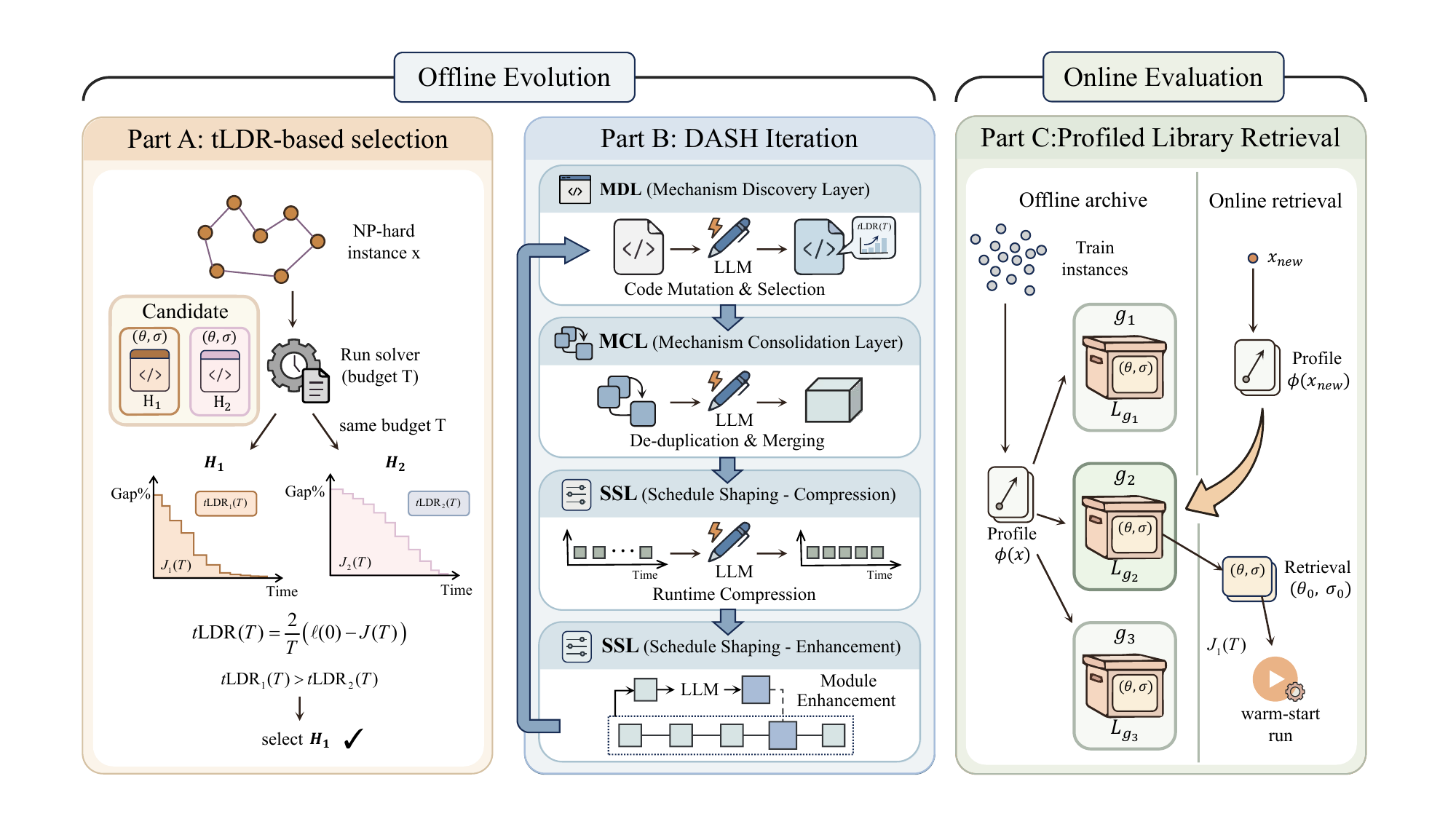}}
   \caption{\textbf{Overview of the DASH framework.}
Offline evolution co-evolves the solver across MDL, MCL, and SSL using terminal log-residual and trajectory efficiency for selection. In parallel, PLR maintains group-wise archives from evaluated candidates.
At test time, PLR retrieves a group-specific solver to warm-start evaluation.}

    \label{mainflow}
\end{figure*}

\subsubsection{Schedule Shaping Layer (SSL)}\label{sec:ssl}

SSL updates the schedule $\sigma$ while keeping the mechanism $\theta$ fixed.
The $\sigma$ specifies how the solver runs over the time budget: which modules are invoked, in what order, and with what triggering rules and per-phase budgets.
In each iteration, we provide the LLM with the parent schedule together with its evaluation traces (e.g., per-module runtime and the resulting convergence trajectory), and the LLM proposes a revised schedule $\sigma'$ by adjusting module allocation and control parameters.
We then evaluate $\sigma'$ under the same protocol as its parent. SSL proceeds in two consecutive stages: it first performs \emph{Compression} to reduce wasted computational slack, and then applies \emph{Enhancement} to spend the recovered budget more effectively.

\paragraph{Stage 1: Compression}
This stage targets computational slack in the execution chain.
Given the parent trace, the LLM revises $\sigma$ by shortening or removing low-efficiency phases and by retuning iteration limits or trigger conditions, so that the solver achieves a comparable batch-averaged terminal log-residual with lower batch-averaged runtime. 
However, $\mathrm{tLDR}(T)$ is typically compared under the same time budget $T$. Once compression changes the realized runtime, trajectory comparisons based on $\mathrm{tLDR}(T)$ become mismatched. We therefore use batch-averaged runtime $\bar{t}$ for schedule-level acceptance in this stage.
A compressed schedule is accepted only if it reduces the batch-averaged runtime beyond the comparison margin while keeping the batch-averaged terminal log-residual within the comparison margin of the parent.

\begin{equation}
\mathrm{Accept}(\sigma') \iff
\big(\bar{t}' \le (1-\epsilon)\bar{t}\big)
\ \land\
\big(|\bar{\ell}'-\bar{\ell}|\le \epsilon\big).
\label{eq:ssl1_rule}
\end{equation}

\paragraph{Stage 2: Enhancement}
Once the schedule is compressed, this stage targets marginal gains in terminal log-residual by further modifying $\sigma$.
Starting from the compressed schedule, the LLM revises $\sigma$ by reallocating the compressed runtime budget across phases and adjusting runtime schedule controls (e.g., module allocation and their triggering rules or iteration limits), so that more of the runtime is spent on phases that are most helpful for improving the incumbent (e.g., a more thorough perturbation or local-improvement setting). An enhanced schedule is accepted only if it improves the batch-averaged terminal log-residual $\bar{\ell}$ beyond the comparison margin and keeps the batch-averaged trajectory efficiency $\bar{k}$ within the comparison margin of the parent.

\begin{equation}
\mathrm{Accept}(\sigma') \iff
\big(\bar{\ell}' \le \bar{\ell} - \epsilon\big)
\ \land\
\big(|\bar{k}'-\bar{k}|\le \epsilon\cdot|\bar{k}|\big).
\label{eq:ssl2_rule}
\end{equation}

To ground the decomposition of solver edits in a concrete TSP backbone, Table~\ref{tab:gls_variants_corrected_tkde} instantiates this decomposition on GLS for TSP by separating editable components into the search mechanism $\theta$ and the runtime schedule $\sigma$. The former specifies the search rules and operators applied by the solver, whereas the latter specifies their activation, frequency, and budget allocation. Accordingly, MDL and MCL revise $\theta$, while SSL edits $\sigma$ through runtime reallocation and control adjustments.

\section{DASH Framework}\label{sec:oplib_retrieval}

DASH co-evolves the solver mechanism $\theta$ and runtime schedule $\sigma$ via iterative LLM-Driven edits across MDL, MCL, and SSL, and introduces Profiled Library Retrieval (PLR) to decouple archiving from population evolution within a single evolutionary process.
We maintain a global population $\mathcal{P}$ of top-$k$ solvers ranked by performance aggregated over all groups and a group-wise archive $L_g$ that stores the top-$k$ specialized solvers for each group $g$.
The global population and the group-wise archives serve different objectives in DASH. The global population drives continued evolution toward a solver that remains competitive on average, whereas the group-wise archives preserve specialists that may be suboptimal globally but particularly effective for specific profile regions.

During evolution, parents are sampled from $\mathcal{P}$. At test time, PLR selects a solver from $L_{g^\star}$ for profile-aware warm starts.
This separation avoids re-evolving a solver for each group and prevents a single archive from being dominated by either global ranking or group specialization in a given batch.
For fairness, each evolutionary run starts from the same task-specific initial solver. Figure~\ref{mainflow} illustrates the end-to-end DASH workflow.

\subsection{Instance Profiles and Instance Groups}

To improve generalization across heterogeneous instances, we construct an offline training set $\mathcal{D}_{\mathrm{train}}$ of randomly generated instances with diverse distributions. For each instance $x$, we compute a lightweight instance profile $\phi(x)\in\mathbb{R}^p$ summarizing key size and structural statistics. We then partition $\mathcal{D}_{\mathrm{train}}$ into $G$ groups $\{\mathcal{D}_g\}_{g=1}^G$ based on profile similarity, each represented by a prototype $\phi_g$.
For routing tasks, these profiles combine scale, distance statistics, and spatial-structure descriptors; for resource-allocation tasks, they summarize value statistics, weight statistics, and capacity tightness across constraints.
Within DASH, they are used both to support stratified batch construction during evolution and to index group-specialized solvers for retrieval during evaluation.

\subsection{Evaluation and Decoupled Archiving}

We execute the iterative DASH loop on $\mathcal{D}_{\mathrm{train}}$ to co-evolve the solver. To efficiently obtain both global and group-specialized solvers, we employ a decoupled archiving strategy. In each iteration, we evaluate candidates on a sampled batch $\mathcal{B}$, composed of batches $\mathcal{B}_g$ sampled from each group $\mathcal{D}_g$:
\begin{equation}
\mathcal{B}= \cup_{g=1}^{G}\mathcal{B}_g,\quad \mathcal{B}_g = \text{Sample}(\mathcal{D}_g,m).
\end{equation}
Each iteration selects a parent solver $\pi=(\theta,\sigma)$ and evaluates the parent together with all candidate solvers on the same union batch. MDL and MCL revise the mechanism $\theta$ while keeping the schedule $\sigma$ fixed, whereas SSL-1 and SSL-2 revise $\sigma$ while keeping $\theta$ fixed.
Based on these evaluations, we (i) update the size-$k$ global population using layer-wise acceptance on batch means, replacing the lowest-ranked solver if accepted, and (ii) update each group archive $L_g$ (top-$k$) using the candidate's batch-mean performance on $\mathcal{B}_g$.
Archive updates are independent of population acceptance: every evaluated candidate can be inserted into $L_g$ if it outperforms the lowest-ranked entry, keeping only the top-$k$ solvers. Within each archive, candidates are ranked primarily by the recorded group-wise terminal log-residual, with trajectory efficiency and runtime used as secondary criteria. This prioritizes solvers that achieve both strong final quality and efficient convergence under the same evaluation protocol. The decoupling allows a candidate to contribute differently to the two objectives of DASH. A solver that does not remain in the global population may still be preserved if it performs strongly on a particular profile group, whereas a solver with stronger average performance can continue to drive evolution even if it is not the best specialist for every group.

\subsection{Online Profiled Library Retrieval (PLR)}

Given a query instance $x_{\mathrm{new}}$, we compute and normalize its instance profile $\phi(x_{\mathrm{new}})$ using the training set statistics. We then identify the most similar group $g^\star$ by minimizing the Euclidean distance to the prototype profiles:
\begin{equation}
g^\star = \arg\min_{g\in\{1,\dots,G\}} \big\| \phi(x_{\mathrm{new}}) - \phi_g \big\|_2,
\label{eq:nearest_group}
\end{equation}
where $\phi_g$ is the prototype profile of group $g$.
Once the best-matching group is identified, we retrieve the best solver $(\theta^\star,\sigma^\star)$ from the corresponding group archive $L_{g^\star}$ and warm-start the new instance with a solver specialized to similar profile statistics instead of restarting online re-adaptation.

\subsection{DASH configuration.} We use GPT-5.4-mini~\cite{openai2025gpt5systemcard} with temperature 0.7 as the base LLM for DASH. We set the shared comparison margin $\epsilon$ to 0.05 in all experiments. Unless otherwise noted, DASH uses a global population size of 5 and a per-group archive size of 5, and performs 100 solver evaluations during offline evolution.

\begin{table*}[t]
\centering
\small
\setlength{\tabcolsep}{0.4pt}
\renewcommand{\arraystretch}{1.05}
\caption{\textbf{Performance comparison on TSP.} We report three metrics: Obj, Gap (\%), and Time (s), where Obj denotes the achieved objective value, Gap (\%) denotes the percentage difference from the task-specific reference solution, and Time (s) denotes the measured wall-clock runtime per instance. All TSP settings use a 10 s runtime budget per instance. Bold and underlined values indicate the best and second-best LHD results.}
\label{tab:tsp20_1000}
\begin{tabular*}{\textwidth}{@{\extracolsep{\fill}}l*{4}{ccc}@{}}
\toprule
& \multicolumn{3}{c}{TSP 20} & \multicolumn{3}{c}{TSP 100} & \multicolumn{3}{c}{TSP 200} & \multicolumn{3}{c}{TSP 500} \\
\cmidrule(lr){2-4}\cmidrule(lr){5-7}\cmidrule(lr){8-10}\cmidrule(lr){11-13}
Method & Obj & Gap (\%) & Time (s) & Obj & Gap (\%) & Time (s) & Obj & Gap (\%) & Time (s) & Obj & Gap (\%) & Time (s) \\
\midrule
\rowcolor{gray!20}\multicolumn{13}{l}{\textbf{Conventional Solver}} \\
Concorde & 3.827 & 0.000 & 0.011 & 7.763 & 0.000 & 0.235 & 10.652 & 0.000 & 1.439 & 16.553 & 0.000 & 7.315 \\
LKH3 & 3.827 & 0.000 & 0.021 & 7.763 & 0.012 & 0.622 & 10.652 & 0.000 & 1.984 & 16.557 & 0.022 & 7.784 \\
OR-Tools & 3.827 & 0.000 & 3.002 & 7.922 & 2.054 & 10.002 & 10.992 & 3.198 & 10.004 & 17.473 & 5.558 & 10.012 \\
\addlinespace[2pt]
\rowcolor{gray!20}\multicolumn{13}{l}{\textbf{Heuristic Algorithms}} \\
LS & 3.834 & 0.199 & 0.003 & 7.892 & 1.666 & 5.301 & 10.982 & 3.107 & 10.000 & 17.843 & 7.794 & 10.067 \\
GLS & 3.827 & 0.000 & 0.095 & 7.908 & 1.877 & 9.861 & 10.949 & 2.795 & 7.067 & 17.612 & 6.400 & 10.114 \\
KGLS & 3.827 & 0.000 & 0.098 & 7.773 & 0.136 & 9.370 & 10.870 & 2.053 & 10.000 & 17.470 & 5.541 & 10.002 \\
ILS & 3.827 & 0.000 & 1.021 & 7.819 & 0.722 & 10.013 & 10.928 & 2.604 & 10.020 & 17.664 & 6.711 & 10.195 \\
LKH & 3.828 & 0.022 & 0.037 & 7.841 & 1.005 & 10.010 & 10.993 & 3.208 & 10.041 & 17.692 & 6.880 & 10.225 \\
\addlinespace[2pt]
\rowcolor{gray!20}\multicolumn{13}{l}{\textbf{NCO}} \\
AM & 3.832 & 0.132 & 0.076 & 8.101 & 4.357 & 0.112 & 11.403 & 7.052 & 9.850 & 17.119 & 3.421 & 10.036 \\
POMO & 3.827 & 0.000 & 0.045 & 7.766 & 0.042 & 0.098 & 10.777 & 1.180 & 9.917 & 17.688 & 6.859 & 9.958 \\
SymNCO & 3.827 & 0.000 & 0.043 & 7.765 & 0.035 & 0.082 & 10.804 & 1.427 & 9.889 & 17.440 & 5.358 & 9.947 \\
DeepACO & 3.827 & 0.009 & 0.269 & 7.766 & 0.043 & 1.806 & 10.715 & 0.594 & 2.589 & 16.863 & 1.871 & 10.220 \\
SIL & 3.827 & 0.000 & 4.489 & 7.824 & 0.794 & 8.940 & 10.763 & 1.049 & 10.000 & 16.954 & 2.423 & 10.000 \\
\addlinespace[2pt]
\rowcolor{gray!20}\multicolumn{13}{l}{\textbf{LHD Frameworks}} \\
FunSearch & \textbf{3.827} & \textbf{0.000} & 0.909 & 7.780 & 0.225 & 10.212 & \underline{10.803} & \underline{1.423} & 10.084 & 16.897 & 2.080 & 10.095 \\
ReEvo& \textbf{3.827} & \textbf{0.000} & 0.763 & 7.777 & 0.191 & 10.095 & 10.857 & 1.925 & \underline{10.044} & 16.891 & 2.040 & 10.167 \\
EoH & \textbf{3.827} & \textbf{0.000} & 1.031 & \underline{7.772} & \underline{0.122} & 10.847 & 10.843 & 1.801 & 10.058 & 16.948 & 2.387 & 10.086 \\
MEoH & \textbf{3.827} & \textbf{0.000} & \underline{0.619} & 7.797 & 0.443 & \underline{9.884} & 10.881 & 2.152 & 10.077 & 17.081 & 3.188 & \underline{9.931} \\
Hercules & \textbf{3.827} & \textbf{0.000} & 0.823 & 7.776 & 0.169 & 10.970 & 10.827 & 1.651 & 10.101 & \underline{16.873} & \underline{1.935} & 10.684 \\
DASH (Ours) & \textbf{3.827} & \textbf{0.000} & \textbf{0.044} & \textbf{7.769} & \textbf{0.086} & \textbf{1.136} & \textbf{10.677} & \textbf{0.243} & \textbf{1.940} & \textbf{16.714} & \textbf{0.974} & \textbf{3.680} \\
\bottomrule
\end{tabular*}
\end{table*}

\section{Experiment}

\subsection{Experimental Protocol}
\label{sec:experimental_protocol}

\noindent\textbf{Datasets and task settings.}
We evaluate DASH on four combinatorial optimization tasks: TSP~\cite{DBLP:conf/icml/0044TY0LWL024}, CVRP~\cite{uchoa2017cvrplib}, VRPTW~\cite{solomon1987vrptw}, and MKP~\cite{Beasley1990orlib}. For TSP, we generate synthetic Euclidean instances with $n\in\{20,100,200,500\}$ nodes by sampling node coordinates independently from $[0,1]$ and additionally use TSPLIB as an external benchmark for transfer evaluation. For CVRP, we generate synthetic Euclidean instances with 100 and 500 customers. For VRPTW, we generate synthetic instances with 15 and 20 customers. For MKP, we use OR-Library~\cite{Beasley1990orlib} benchmark instances with 500 items and 10 and 30 constraints.

\noindent\textbf{Training and test construction.}
For each task, DASH evolves solvers on a task-specific evolution set and reports final performance on disjoint test instances. For TSP, CVRP, and VRPTW, we generate task-specific synthetic instance pools and split them into an evolution set $\mathcal{D}_{\mathrm{train}}$ and a held-out test set $\mathcal{D}_{\mathrm{test}}$. The evolution set is used for instance profiling, K-means grouping into $G{=}10$ groups, and per-iteration batch sampling with $m{=}3$ instances per group ($|\mathcal{B}|{=}30$), whereas the held-out test set is used only for final reporting. For TSP, TSPLIB is used only for external transfer evaluation. For MKP, DASH is evolved on a separate synthetic training set under the same grouping and batch-construction protocol, while final results are reported only on the selected OR-Library benchmark instances. These benchmark instances are never reused for solver evolution, archive updates, or PLR updates.

\noindent\textbf{Evaluation metrics and reference solutions.}
We report \textit{Gap} and \textit{Time}, where \textit{Gap} measures the relative difference to a task-specific reference value and \textit{Time} is the runtime per instance under the specified budget. Lower gap is better for all tasks. The reference value is task-specific: for synthetic TSP, we run Concorde and use the mean objective value over five runs as the reference; for synthetic CVRP and VRPTW, we run PyVRP and use the mean objective value over five runs as the reference; for MKP, we use the best-known values provided by OR-Library, and the reported gap measures the shortfall from these values.

\noindent\textbf{Shared solver backbones.} For TSP, we use Guided Local Search (GLS)~\cite{voudouris1999gls} as the primary backbone, and additionally instantiate DASH on Iterated Local Search (ILS)~\cite{DBLP:books/sp/03/LourencoMS03} and a Python implementation of LKH~\cite{helsgaun2000lkh}. For CVRP, VRPTW, and MKP, we use Ant Colony Optimization (ACO)~\cite{dorigo2004aco} as the shared backbone.

\noindent\textbf{Compared Baselines.} We compare (i) dedicated solvers (Concorde~\cite{applegate2006tsp}, LKH3~\cite{helsgaun_lkh3}, OR-Tools~\cite{ortools}), (ii) handcrafted heuristics (LS~\cite{aarts2003localsearch}, GLS~\cite{voudouris1999gls}, KGLS~\cite{arnold2019kgls}), (iii) neural combinatorial optimization (NCO) methods (Attention Model (AM)~\cite{am_kool2019}, POMO~\cite{pomo_kwon2020}, Sym-NCO~\cite{symnco_kim2022}, DeepACO~\cite{deepaco_ye2023}, SIL~\cite{luo2025boosting}), and (iv) LHD frameworks (FunSearch~\cite{romera2024funsearch}, ReEvo~\cite{DBLP:conf/nips/Ye0CBHKPS24}, EoH~\cite{DBLP:conf/icml/0044TY0LWL024}, MEoH~\cite{yao2025multi}, Hercules~\cite{wu2025efficient}). All LHD baselines use GPT-5.4-mini~\cite{openai2025gpt5systemcard}, are evaluated under the same task-specific solver budget and wall-clock accounting as DASH within each task, and perform 100 solver evaluations during offline evolution. To preserve fairness while keeping each framework operational, we retain each baseline's native search workflow whenever it remains compatible with the shared backbone. Accordingly, baseline edits are restricted to mechanism-side heuristic logic on the shared solver skeleton. For CVRP, VRPTW, and MKP, this editable scope covers construction, pheromone, and repair or local-improvement rules under a fixed schedule. DASH additionally optimizes schedule-side controls on top of mechanism updates.

\begin{table*}[t]
\centering
\scriptsize
\setlength{\tabcolsep}{3pt}
\renewcommand{\arraystretch}{1}
\caption{\textbf{Performance comparison on CVRP/VRPTW/MKP.} We report two metrics: Gap (\%) and Time (s). CVRP 100 and CVRP 500 denote CVRP instances with 100 and 500 customers and use a 5 s runtime budget. VRPTW 15 and VRPTW 20 denote VRPTW instances with 15 and 20 customers and use a 60 s runtime budget. MKP 10 and MKP 30 denote MKP instances with 500 items and 10 and 30 constraints and use a 5 s runtime budget. Bold and underlined values indicate the best and second-best LHD results.}
\label{tab:aco_pred_refined}
\resizebox{\linewidth}{!}{%
\begin{tabular}{l*{12}{c}}
\toprule
Method
& \multicolumn{2}{c}{CVRP 100}
& \multicolumn{2}{c}{CVRP 500}
& \multicolumn{2}{c}{VRPTW 15}
& \multicolumn{2}{c}{VRPTW 20}
& \multicolumn{2}{c}{MKP 10}
& \multicolumn{2}{c}{MKP 30} \\
\cmidrule(lr){2-3}\cmidrule(lr){4-5}\cmidrule(lr){6-7}\cmidrule(lr){8-9}\cmidrule(lr){10-11}\cmidrule(lr){12-13}
& Gap (\%) & Time (s)
& Gap (\%) & Time (s)
& Gap (\%) & Time (s)
& Gap (\%) & Time (s)
& Gap (\%) & Time (s)
& Gap (\%) & Time (s) \\
\midrule
FunSearch & 0.592 & 5.043 & 6.487 & 5.084 & 2.437 & \underline{47.538} & 5.684 & \underline{53.417} & \textbf{0.964} & 5.083 & \textbf{1.861} & 5.106 \\
EoH       & 0.458 & 5.021 & \textbf{3.742} & 5.086 & 4.126 & 56.241 & 7.836 & 58.274 & 1.781 & 5.012 & 2.736 & 5.018 \\
ReEvo     & 0.447 & 5.036 & 5.286 & 5.093 & 2.781 & 59.826 & 5.143 & 54.612 & 1.327 & 4.742 & 2.214 & 4.893 \\
MEoH      & 0.682 & \underline{5.017} & 6.018 & \underline{5.074} & \textbf{1.918} & 60.127 & \textbf{4.638} & 60.356 & 1.546 & \underline{4.618} & 2.438 & \underline{4.781} \\
Hercules  & \underline{0.431} & 5.074 & 5.167 & 5.081 & 3.347 & 60.642 & 6.913 & 60.781 & 2.084 & 5.241 & 3.147 & 5.318 \\
DASH (Ours)     & \textbf{0.334} & \textbf{2.013}
          & \underline{3.935} & \textbf{3.442}
          & \underline{2.084} & \textbf{18.731}
          & \underline{4.927} & \textbf{22.614}
          & \underline{1.083} & \textbf{2.684}
          & \underline{1.978} & \textbf{2.941} \\
\bottomrule
\end{tabular}%
}
\end{table*}

\begin{figure*}[t]
    \centering
\centerline{\includegraphics[width=\linewidth]{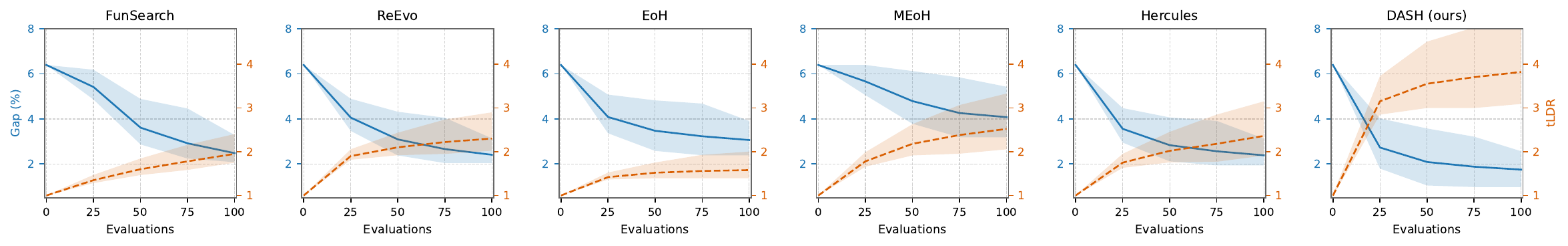}}
\caption{\textbf{Evolutionary dynamics of major LHD frameworks on TSP500.}
Across 5 independent runs (100 evaluations each), we report the best-so-far gap (blue, left axis) and tLDR (orange, right axis) for FunSearch, ReEvo, EoH, MEoH, Hercules, and DASH. Lines show the mean and shaded bands show the variability across runs.}
    \label{fig:mainflow_dynamics}
\end{figure*}

Our experiments are designed as follows:
\begin{itemize}[noitemsep]
    \item \textbf{RQ1 (Main Results):} How does DASH compare with prior LHD frameworks in solution quality and runtime efficiency across combinatorial optimization tasks?
    \item \textbf{RQ2 (Component and Design Analysis):} How do the components and design choices of DASH contribute to solution quality and runtime efficiency?
    \item \textbf{RQ3 (Generalizability and Robustness Analysis):} How well does DASH transfer across solver backbones and distribution shifts?
    \item \textbf{RQ4 (Profile-Aware Retrieval Analysis):} How does profile-aware retrieval organize heterogeneous instances, and how does grouping granularity affect specialization and efficiency?
\end{itemize}

\subsection{Main Results (RQ1)}
\label{sec:exp_main_results}

Table~\ref{tab:tsp20_1000} reports the main TSP results across four scales. Overall, DASH achieves the best balance between gap and time, and its advantage becomes most evident on larger instances. We draw the following observations:
\begin{itemize}[leftmargin=*,noitemsep]
\item DASH is consistently competitive with strong conventional, neural, and prior LHD baselines. On the hardest setting TSP500, it improves the GLS backbone from 6.400\% gap at 10.114 s to 0.974\% gap at 3.680 s, validating the benefit of jointly optimizing mechanisms and schedules.
\item The advantage of DASH appears across scales. On TSP20 and TSP100, some LHD baselines already reach zero or near-zero gap, but further evolution does not reduce the final gap and can even increase runtime, especially on TSP20. In contrast, DASH can still reduce runtime while preserving the same final gap, showing that runtime schedule can improve efficiency even when solution quality is already near-optimal.
\item Compared with prior endpoint-driven LHD baselines such as EoH, DASH achieves lower runtime together with lower final gap on larger instances. On TSP500, DASH reaches a 0.974\% gap in 3.680 s, whereas the compared baselines remain close to the full budget and still produce gaps around 2\% or higher, which is consistent with the effect of trajectory-aware selection and runtime schedule optimization in DASH.
\item Compared with the efficiency-aware LHD baseline MEoH, DASH maintains better gap while preserving a runtime advantage on larger instances. MEoH reduces runtime more efficiently on smaller instances, but on TSP500 its final gap remains clearly higher than that of DASH. This shows that runtime reduction alone does not lead to better overall performance on larger instances.
\end{itemize}

Table~\ref{tab:aco_pred_refined} further validates the effectiveness of DASH on other combinatorial optimization tasks. We consider two related routing tasks, CVRP and VRPTW, and extend the evaluation to a different resource-allocation task, MKP. To ensure a fair comparison, we use ACO as the shared backbone and keep the editable solver scope the same for all LHD baselines, while DASH additionally optimizes the runtime schedule. 
For CVRP, which extends TSP with capacity constraints, DASH achieves the lowest gap on CVRP100 while reducing runtime. On CVRP500, it remains close to the stronger baselines in gap while reducing runtime from about 5 s to 3.442 s.
VRPTW increases the difficulty by introducing time-window constraints, and we accordingly use a larger runtime budget. Under this setting, DASH reduces runtime while keeping the final gap in a comparable range. 
We further extend the evaluation to MKP. On both MKP10 and MKP30, DASH reduces runtime from about 5 s to around 3 s while keeping the final gap close to the stronger baselines.

Figure~\ref{fig:mainflow_dynamics} visualizes the evolutionary dynamics on TSP500 across repeated runs. FunSearch, ReEvo, EoH, and Hercules all reduce the best-so-far gap over evaluations, but their tLDR curves remain lower and increase more gradually. MEoH also raises tLDR more quickly, but its gap curve flattens earlier. Since DASH uses tLDR as the trajectory-aware selection signal, the retained solvers show both faster tLDR growth and stronger best-so-far gap reduction over evaluations.

\subsection{Component and Design Analysis (RQ2)}
\label{sec:ablation}

We use TSP100 and CVRP100 as representative tasks for component and design analysis. Table~\ref{tab:ablation_tsp_cvrp} validates the contribution of DASH's components. Removing SSL-1 causes the largest slowdown on both tasks, indicating that most runtime reduction comes from schedule compression. Removing tLDR from selection degrades both runtime and solution quality, consistent with its role in favoring solvers with efficient early convergence. In contrast, removing SSL-2 mainly degrades the final gap with only minor runtime changes, matching its role as a quality-oriented enhancement stage. Finally, disabling PLR worsens both gap and time, showing that profile-based retrieval improves group-specific solver selection at inference.

\begin{table}[h]
\centering
\small
\setlength{\tabcolsep}{4.2pt}
\renewcommand{\arraystretch}{1}
\caption{\textbf{Ablation study on DASH components.} We report Gap (\%) and Time (s) on TSP100 and CVRP100.}
\label{tab:ablation_tsp_cvrp}
\begin{tabular*}{\columnwidth}{@{\extracolsep{\fill}}lcccc@{}}
\toprule
\multirow{2}{*}{Variant} &
\multicolumn{2}{c}{TSP 100} &
\multicolumn{2}{c}{CVRP 100} \\
\cmidrule(lr){2-3}\cmidrule(lr){4-5}
& Gap (\%) & Time (s) & Gap (\%) & Time (s) \\
\midrule
\textbf{DASH (Full)} & \textbf{0.086} & \textbf{1.136} & \textbf{0.334} & \textbf{2.013} \\
w/o tLDR & 0.452 & 2.200 & 0.385 & 4.859 \\
w/o SSL-1 & 0.105 & 8.505 & 0.362 & 6.201 \\
w/o SSL-2 & 0.165 & 1.002 & 0.392 & 1.920 \\
w/o PLR & 0.155 & 1.980 & 0.450 & 2.963 \\
\bottomrule
\end{tabular*}
\end{table}

Table~\ref{tab:layer_ablation} further separates the roles of the three iteration layers. Removing MDL causes the largest gap degradation, indicating that mechanism discovery is the main source of the gap reduction. Removing MCL mainly increases runtime while only moderately affecting gap, which is consistent with MCL pruning redundant branches without materially changing the schedule. Removing SSL almost eliminates the efficiency advantage of DASH, showing that runtime schedule optimization is the primary driver of runtime reduction.

\begin{table}[h]
\centering
\small
\renewcommand{\arraystretch}{1}
\caption{\textbf{Ablation study on DASH iteration layers.} We report Gap (\%) and Time (s) on TSP100 and CVRP100.}
\label{tab:layer_ablation}
\begin{tabular*}{\columnwidth}{@{\extracolsep{\fill}}lcccc@{}}
    \toprule
    \multirow{2}{*}{Variant} &
    \multicolumn{2}{c}{TSP 100} &
    \multicolumn{2}{c}{CVRP 100} \\
    \cmidrule(lr){2-3}\cmidrule(lr){4-5}
    & Gap (\%) & Time (s) & Gap (\%) & Time (s) \\
    \midrule
    \textbf{DASH (Full)} & \textbf{0.086} & \textbf{1.136} & \textbf{0.334} & \textbf{2.013} \\
    w/o MDL & 0.625 & 0.950 & 0.812 & 1.845 \\
    w/o MCL & 0.114 & 1.680 & 0.358 & 2.550 \\
    w/o SSL (Full) & 0.158 & 9.105 & 0.380 & 8.950 \\
    \bottomrule
\end{tabular*}
\end{table}

We also examine how the trajectory metric used in selection affects the resulting gap and runtime. Concretely, we replace tLDR with an alternative metric while keeping the rest of DASH unchanged. We compare tLDR against three alternatives: \textbf{Terminal Time}, which uses runtime $t_{\mathrm{run}}$ as the efficiency signal; \textbf{Time-to-10\%}, the time required to reach $0.1\cdot \mathrm{Gap}(0)$ on the incumbent trajectory, capped at $T$ if the threshold is not reached; and \textbf{Linear AUC}, the normalized linear-space area under the best-so-far gap curve, $\frac{1}{T}\int_0^T \mathrm{Gap}_{\mathrm{best}}(\tau)\,d\tau$.


\begin{table}[h]
\centering
\small
\setlength{\tabcolsep}{4.2pt}
\renewcommand{\arraystretch}{1}
\caption{\textbf{Sensitivity to selection design.}
We study two design choices in DASH: the trajectory metric used in selection and the shared comparison margin $\epsilon$. Results are reported on TSP 100 and CVRP 100.}
\label{tab:selection_sensitivity}
\begin{tabular*}{\columnwidth}{@{\extracolsep{\fill}}lcccc@{}}
\toprule
\multirow{2}{*}{Variant}
& \multicolumn{2}{c}{TSP 100}
& \multicolumn{2}{c}{CVRP 100} \\
\cmidrule(lr){2-3}\cmidrule(lr){4-5}
& Gap (\%) & Time (s) & Gap (\%) & Time (s) \\
\midrule
\rowcolor{gray!20}\multicolumn{5}{l}{\hspace{-\tabcolsep}\textbf{Trajectory Metric for Selection}} \\
DASH (tLDR) & \textbf{0.086} & 1.136 & \textbf{0.334} & 2.013 \\
Terminal Time & 0.585 & \textbf{0.850} & 0.852 & \textbf{1.420} \\
Time-to-10\% & 0.242 & 1.020 & 0.425 & 1.880 \\
Linear AUC & 0.130 & 1.105 & 0.468 & 1.950 \\
\addlinespace[2pt]
\rowcolor{gray!20}\multicolumn{5}{l}{\hspace{-\tabcolsep}\textbf{Comparison Margin}} \\
$\epsilon = 0.01$ & 0.118 & \textbf{1.028} & 0.368 & \textbf{1.926} \\
$\epsilon = 0.05$ & \textbf{0.086} & 1.136 & \textbf{0.334} & 2.013 \\
$\epsilon = 0.10$ & 0.147 & 1.284 & 0.401 & 2.467 \\
\bottomrule
\end{tabular*}
\end{table}

Replacing tLDR with alternative metrics changes the gap and time trade-off in different ways. Using \textit{Terminal Time} yields the lowest runtime but also the worst gap on both tasks, indicating that a purely runtime-oriented signal tends to select solvers with weaker refinement. \textit{Time-to-10\%} and \textit{Linear AUC} recover part of this loss, but both remain inferior to tLDR. \textit{Time-to-10\%} captures whether a solver reaches an early target quickly, but it does not distinguish candidates well after the threshold is reached. \textit{Linear AUC} uses the full trajectory, but in linear space it is less sensitive to later-stage improvements when the gap is already small. In contrast, tLDR achieves the best gap with only moderate runtime increase, which is consistent with using the full convergence trajectory in log-space as the selection signal rather than relying only on runtime, a single threshold, or a linear-space summary.

Table~\ref{tab:selection_sensitivity} further examines how the trade-off between gap and runtime changes with the shared comparison margin $\epsilon$. In our experiments, $\epsilon=0.01$ makes acceptance more aggressive and slightly lowers runtime, but leads to worse final gap on both tasks, while $\epsilon=0.10$ yields more conservative selection and tends to reject moderate but still useful improvements, resulting in a weaker balance between gap and runtime. We therefore use $\epsilon=0.05$ as a practical default in the experiments.

\begin{table}[h]
\centering
\small
\setlength{\tabcolsep}{5.2pt}
\renewcommand{\arraystretch}{1}
\caption{\textbf{Transferability of DASH across solver frameworks on TSPLIB.}
We report Gap (\%) and Time (s) on small (14--200) and large (201--1000) instances.}
\label{tab:tsp_transfer_framework}
\begin{tabular*}{\columnwidth}{@{\extracolsep{\fill}}lcccc@{}}
\toprule
\multirow{2}{*}{Method}
& \multicolumn{2}{c}{TSPLIB-small}
& \multicolumn{2}{c}{TSPLIB-large} \\
\cmidrule(lr){2-3}\cmidrule(lr){4-5}
& Gap (\%) & Time (s) & Gap (\%) & Time (s) \\
\midrule
GLS              & 1.263 & 1.605 & 4.581 & 32.597 \\
ILS              & 1.209 & 2.231 & 3.969 & 38.212 \\
LKH              & 0.957 & 3.777 & 3.718 & 33.001 \\
\midrule
DASH+GLS         & 0.122 & 0.263 & 1.224 & 4.520 \\
DASH+ILS         & 0.102 & 0.276 & 1.157 & 5.300 \\
DASH+LKH         & 0.092 & 0.315 & 1.070 & 4.297 \\
\bottomrule
\end{tabular*}
\end{table}

\subsection{Generalizability and Robustness Analysis (RQ3)}
\label{sec:generalizability}

Besides validating DASH across different tasks, we also examine whether its evolutionary optimization transfers across different solver backbones within the same task. 
Table~\ref{tab:tsp_transfer_framework} shows that DASH consistently improves all three TSP backbones (GLS/ILS/LKH) on TSPLIB, reducing both gap and runtime on small and large instances. The gains across distinct backbones indicate that DASH acts as a general optimizer rather than being tied to a specific solver.

\begin{figure*}[h]
  \centering
\centerline{\includegraphics[width=\linewidth]{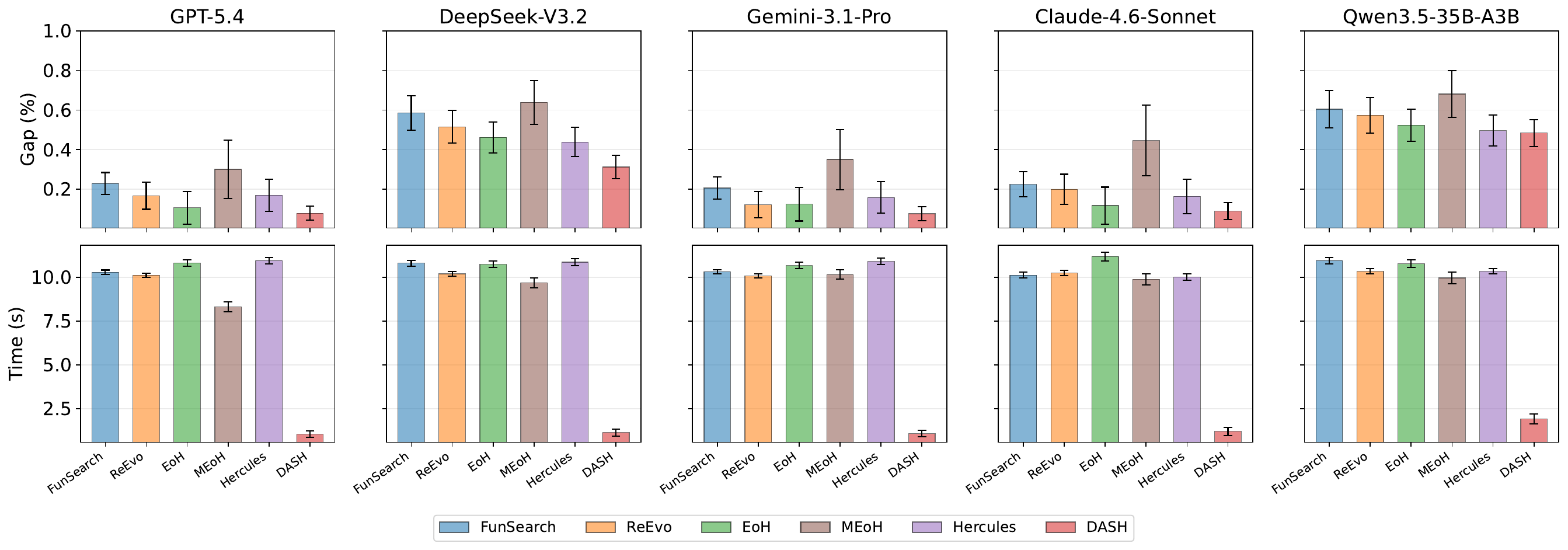}}
\caption{\textbf{Base-model sensitivity on TSP 100.}
Using five base LLMs for solver generation (5 runs, 100 evaluations each), we report the final best-so-far gap (top) and the runtime of that best solver (bottom).}
    \label{fig:basemodel_sensitivity}
\end{figure*}

\begin{table}[h]
\centering
\scriptsize
\setlength{\tabcolsep}{2.6pt}
\renewcommand{\arraystretch}{1}
\caption{\textbf{Generalization Cost on TSPLIB (14--1000) with 100 evaluations.} (+Iters) increases the evaluation to $5\times$ (500 evaluations); (+Groups) applies the same instance grouping protocol as DASH.}
\label{tab:cost_gen_tsp}
\begin{tabular*}{\columnwidth}{@{\extracolsep{\fill}}lccccc@{}}
\toprule
\textbf{Method}
& \multicolumn{2}{c}{\textbf{Evolution}}
& \multicolumn{2}{c}{\textbf{Test}}
& \textbf{Tokens (k)} \\
\cmidrule(lr){2-3}\cmidrule(lr){4-5}
& Gap (\%) & Time (s) & Gap (\%) & Time (s) & Input/Output \\
\midrule
MEoH              & 2.070 & 20.027  & 3.174 & 14.422 & 44.7/27.3 \\
MEoH (+Iters)     & 1.891 & 21.143  & 2.923 & 14.146 & 229.5/131.0 \\
MEoH (+Groups)    & 1.660 & 20.521  & 1.690 & 12.967 & 472.9/280.6 \\
\midrule
Hercules          & 1.354 & 20.283 & 0.867 & 7.789 & 143.4/31.2 \\
Hercules (+Iters) & 1.125 & 19.990 & 0.822 & 7.990 & 711.4/155.7 \\
Hercules (+Groups)& 0.792 & 18.231 & 0.480 & 7.142 & 1401.1/322.3 \\
\midrule
DASH              & \textbf{0.516} & \textbf{4.340}  & \textbf{0.532} & \textbf{1.846} & \textbf{115.0/58.8} \\
\bottomrule
\end{tabular*}
\end{table}

Table~\ref{tab:cost_gen_tsp} studies generalization under a distribution shift by evaluating solvers evolved under the same 100-evaluation setting on TSPLIB. For baselines, (+Iters) increases the evolution budget to 5 $\times$, and (+Groups) trains group-specific solvers using the same grouping protocol as DASH. Among the compared baselines, MEoH uses the fewest tokens in the standard setting, but its gap increases clearly from evolution to TSPLIB. Increasing the evaluation budget or introducing group-specific training can reduce this shift, but both require substantially higher token cost.
Compared with MEoH, Hercules is more stable under the same shift, which is consistent with its use of historical search experience during heuristic generation and evaluation.
In contrast, DASH maintains low gap from evolution to TSPLIB test while also achieving the lowest runtime on both sides, showing that PLR improves robustness to distribution shift without the additional optimization cost.

\begin{table}[t]
\centering
\small
\setlength{\tabcolsep}{5.2pt}
\renewcommand{\arraystretch}{1}
\caption{\textbf{TSPLIB performance under time-constraint evaluation.}
We fix the time budget across all methods, set $T{=}10$s for TSPLIB-small (14--200) and $T{=}60$s for TSPLIB-large (201--1000).}
\label{tab:tsplib_equal_budget_est}
\begin{tabular*}{\columnwidth}{@{\extracolsep{\fill}}lcccc@{}}
\toprule
\multirow{2}{*}{Method}
& \multicolumn{2}{c}{TSPLIB-small}
& \multicolumn{2}{c}{TSPLIB-large} \\
\cmidrule(lr){2-3}\cmidrule(lr){4-5}
& Gap (\%) & Time (s) & Gap (\%) & Time (s) \\
\midrule
\rowcolor{gray!20}\multicolumn{5}{l}{\hspace{-\tabcolsep}\textbf{Conventional Solver}} \\
Concorde & 0.000 & 0.332 & 0.000 & 11.797 \\
LKH3 & 0.000 & 0.211 & 0.000 & 2.789 \\
OR-Tools & 1.291 & 7.310 & 3.586 & 60.000 \\
\addlinespace[2pt]
\rowcolor{gray!20}\multicolumn{5}{l}{\hspace{-\tabcolsep}\textbf{LHD Frameworks}} \\
FunSearch & 1.163 & 1.048 & 2.760 & 12.898 \\
ReEvo   & 1.396 & 2.503 & 3.189 & 36.996 \\
EoH   & 3.162 & 1.574 & 4.021 & 24.225 \\
MEoH   & 3.042 &  1.079 & 3.397 & 36.968 \\
Hercules  & 0.216 & 0.576 & 1.967 & 19.977 \\
DASH  & 0.122 &  0.263 & 1.224 & 4.520 \\
\bottomrule
\end{tabular*}
\end{table}

\begin{table}[t]
\centering
\small
\setlength{\tabcolsep}{5.2pt}
\renewcommand{\arraystretch}{1}
\caption{\textbf{TSPLIB results under time-unconstrained evaluation.}
We report Gap (\%) and Time (s) on TSPLIB-small (14--200) and TSPLIB-large (201--1000). For DASH, SSL is disabled during evaluation.}
\label{tab:tsplib_run_to_stop_all}
\begin{tabular*}{\columnwidth}{@{\extracolsep{\fill}}lcccc@{}}
\toprule
\multirow{2}{*}{Method}
& \multicolumn{2}{c}{TSPLIB-small}
& \multicolumn{2}{c}{TSPLIB-large} \\
\cmidrule(lr){2-3}\cmidrule(lr){4-5}
& Gap (\%) & Time (s) & Gap (\%) & Time (s) \\
\midrule
\rowcolor{gray!20}\multicolumn{5}{l}{\hspace{-\tabcolsep}\textbf{Conventional Solver}} \\
Concorde & 0.000 & 0.273 & 0.000 & 10.285 \\
LKH3 & 0.000 & 0.150 & 0.001 & 5.118 \\
OR-Tools & 2.650 & 0.452 & 4.206 & 15.619 \\
\addlinespace[2pt]
\rowcolor{gray!20}\multicolumn{5}{l}{\hspace{-\tabcolsep}\textbf{LHD Frameworks}} \\
FunSearch & 1.160 &    1.739 & 2.607 & 16.820 \\
ReEvo    & 1.411 &    2.668 & 1.443 & 125.306 \\
EoH    & 3.132 &   0.970 & 4.022 & 26.714 \\
MEoH  & 3.035 &    1.066 & 3.397 & 23.657 \\
Hercules & 0.096 &    0.708 & 1.742 & 20.609 \\
DASH w/o SSL   & 0.010 &    2.104 & 0.824 &   26.240 \\
\bottomrule
\end{tabular*}
\end{table}

Table~\ref{tab:tsplib_equal_budget_est} compares Conventional Solvers and LHD Frameworks under time-constrained evaluation, where the time budget is fixed to 10 s for TSPLIB-small and 60 s for TSPLIB-large. Under the same budget, DASH achieves the best results among the compared LHD frameworks on both instance scales. On TSPLIB-small, DASH surpasses best-performing LHD framework Hercules, reducing the gap from 0.216\% to 0.122\% and the runtime from 0.576 s to 0.263 s. On TSPLIB-large, DASH again achieves both lower gap and lower runtime, improving from 1.967\% to 1.224\% in gap and from 19.977 s to 4.520 s in runtime. 
DASH is also the LHD method closest to the C implementation LKH3, especially on TSPLIB-small where it reaches near-zero gap with comparable runtime.

Table~\ref{tab:tsplib_run_to_stop_all} further compares Conventional Solvers and LHD Frameworks under time-unconstrained evaluation. We remove the time budget and, for LHD frameworks, fix the GLS backbone to 1000 iterations. For a fair comparison, we disable SSL in DASH. 
Under this setting, most LHD frameworks improve further, especially on TSPLIB-large.
ReEvo provides a representative example: its gap on TSPLIB-large improves from 3.189\% to 1.443\%, but its runtime increases from 36.996 s to 125.306 s. Under the same unconstrained setting, DASH w/o SSL reaches 0.824\% on TSPLIB-large in 26.240 s, and 0.010\% on TSPLIB-small in 2.104 s. Although SSL is disabled in this setting, these results still show that the mechanisms evolved by DASH retain clear potential for further improvement when additional solving budget is allowed.

\begin{figure*}[t]
  \centering
  \centerline{\includegraphics[width=\linewidth]{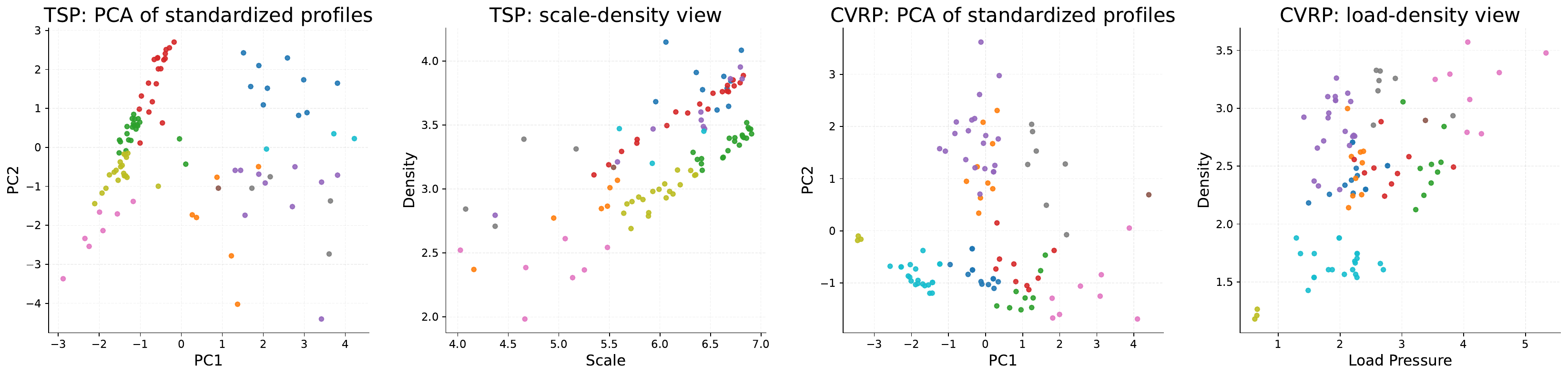}}
\caption{\textbf{Instance profiling and grouping for TSP and CVRP.}
We visualize standardized profile vectors for 100 TSP instances and 100 CVRP instances, and apply k-means with $G{=}10$.
\emph{(a)} PCA projection of TSP profiles, colored by group assignment.
\emph{(b)} TSP scale--density view using the scale feature $\log(n)$ and a nearest-neighbor based density proxy.
\emph{(c)} PCA projection of CVRP profiles, colored by group assignment.
\emph{(d)} CVRP load--density view using $\log(\sum_i d_i/Q)$ and a geometric density proxy.}
  \label{fig:profile_demo_pca}
\end{figure*}

Figure~\ref{fig:basemodel_sensitivity} evaluates sensitivity to the underlying code-generation model. Changing the base LLM does affect the absolute performance of all methods: the overall gaps are lower under GPT-5.4~\cite{openai2025gpt5systemcard}, Gemini-3.1-Pro~\cite{googledeepmind2025gemini3pro_modelcard}, and Claude-4.6-Sonnet~\cite{anthropic2026claudesonnet46}, and become larger under DeepSeek-V3.2~\cite{deepseek_v3_2_2025} and Qwen3.5-35B-A3B~\cite{DBLP:journals/corr/abs-2505-09388}. Nevertheless, DASH remains consistently strong across all five models, maintaining the lowest or near-lowest gap together with a clear runtime advantage. This shows that the advantage of DASH is not tied to a single base model.

\begin{table*}[t]
\centering
\small
\setlength{\tabcolsep}{3.2pt}
\renewcommand{\arraystretch}{1.1}
\caption{\textbf{Sensitivity to the number of instance groups $G$ on TSP and CVRP.}
We vary the grouping granularity $G\in\{5,10,15,20\}$ under the same evolution budget (100 evaluations). We report evolution/test gap and runtime, together with evaluation cost per iteration (\textit{Eval./iter.}) and total evaluation cost (\textit{Total Eval.}).}
\label{tab:sens_group_G}
\begin{tabular*}{\textwidth}{@{\extracolsep{\fill}}c|*{6}{cc}@{}}
\toprule
\multirow{2}{*}{\textbf{$G$}}
& \multicolumn{2}{c}{\textbf{Evo. Gap (\%)}}
& \multicolumn{2}{c}{\textbf{Evo. Time (s)}}
& \multicolumn{2}{c}{\textbf{Test Gap (\%)}}
& \multicolumn{2}{c}{\textbf{Test Time (s)}}
& \multicolumn{2}{c}{\textbf{Eval./iter. (s)}}
& \multicolumn{2}{c}{\textbf{Total Eval. (min)}} \\
\cmidrule(lr){2-3}\cmidrule(lr){4-5}\cmidrule(lr){6-7}\cmidrule(lr){8-9}\cmidrule(lr){10-11}\cmidrule(lr){12-13}
& \textbf{TSP} & \textbf{CVRP}
& \textbf{TSP} & \textbf{CVRP}
& \textbf{TSP} & \textbf{CVRP}
& \textbf{TSP} & \textbf{CVRP}
& \textbf{TSP} & \textbf{CVRP}
& \textbf{TSP} & \textbf{CVRP} \\
\midrule
5
& 0.550 & 0.416
& 2.050 & 2.325
& 0.620 & 0.469
& 1.950 & 1.911
& 28.1 & 31.9
& 46.7 & 53.0 \\
10
& 0.419 & 0.317
& 1.905 & 2.160
& 0.442 & 0.334
& 1.775 & 2.013
& 35.0 & 39.7
& 58.3 & 66.1 \\
15
& 0.410 & 0.310
& 1.915 & 2.172
& 0.438 & 0.331
& 1.780 & 2.019
& 50.5 & 57.3
& 83.3 & 94.5 \\
20
& 0.408 & 0.308
& 1.925 & 2.183
& 0.437 & 0.330
& 1.785 & 2.024
& 65.7 & 74.5
& 108.3 & 122.8 \\
\bottomrule
\end{tabular*}
\end{table*}

\subsection{Profile-Aware Retrieval Analysis (RQ4)}

The TSPLIB transfer results already show that PLR reduces the need to restart adaptation under shifted instance distributions. To support this behavior, DASH constructs a lightweight profile vector for each instance, standardizes these features, and clusters them into $G$ groups. The resulting groups are used both for stratified evolution and for test-time retrieval through the nearest group prototype.

To illustrate the profile structure, we characterize the features of 100 TSP instances and 100 CVRP instances. For TSP, the profiles include scale, distance statistics, local density, and shape descriptors. For CVRP, the profiles include instance scale, geometric density, depot-relative structure, demand variation, and load pressure. Figure~\ref{fig:profile_demo_pca} shows that the resulting groups separate both TSP and CVRP instances along interpretable profile dimensions rather than simply partitioning by size.

Table~\ref{tab:sens_group_G} further studies the sensitivity to the number of instance groups $G$. Increasing $G$ beyond 10 yields marginal gap improvements on both tasks, while evaluation cost rises because the same total evolution budget must be spread across more groups.
We therefore use $G{=}10$ as the default trade-off between specialization and evaluation efficiency.

\section{Conclusion}

We revisit LLM-Driven Heuristic Design from a dynamics perspective, evaluating solvers by their convergence trajectories rather than only terminal gap. We introduce the Trajectory-aware Lyapunov Decay Rate (tLDR), a metric computed from execution traces that captures the rate and consistency of convergence throughout the run. Guided by tLDR, DASH co-evolves search mechanisms and runtime schedules through three iteration layers: MDL, MCL, and SSL under a unified acceptance protocol, so that solver generation is driven not only by final quality but also by how efficiently useful progress is produced over time. The ablation results further show that these components play different and complementary roles: MDL mainly improves the quality of discovered mechanisms, MCL reduces redundant exploration, and SSL converts those stronger candidates into shorter realized runtimes.

To reduce re-adaptation under heterogeneous instance groups, DASH incorporates Profiled Library Retrieval (PLR), which decouples group-specific archiving from global evolution to enable profile-aware warm starts. This makes it possible to harvest specialized solvers during the evolutionary process instead of restarting adaptation from scratch whenever the instance distribution shifts. Across four combinatorial optimization problems, DASH improves runtime efficiency by over 4$\times$ while achieving a better balance between gap and runtime than prior LHD baselines. It also generalizes across multiple solver backbones and maintains performance under distribution shift with substantially lower adaptation cost.

\clearpage
\newpage

\bibliography{custom}
\end{document}